%% file: lrec-coling2024.tex
\documentclass[10pt, a4paper]{article}
\usepackage{amsmath}
\usepackage{amssymb}
\usepackage{pifont}
\usepackage{lrec-coling2024} 

\usepackage{stfloats}
\usepackage{rotating}
\usepackage{makecell}
\usepackage{enumitem}
\usepackage{tabularray}
\usepackage{xcolor}
\usepackage{multirow}
\usepackage{multicol}
\usepackage{pifont}
\UseTblrLibrary{booktabs}

\definecolor{mblue}{HTML}{003F5C}
\definecolor{mviolet}{HTML}{58508D}
\definecolor{mpurple}{HTML}{BC5090}
\definecolor{msalmon}{HTML}{FF6361}
\definecolor{mgold}{HTML}{FFA600}

\definecolor{CarnegieRed}{HTML}{CC002B}
\definecolor{Black}{HTML}{000000}
\definecolor{SteelGray}{HTML}{B3B3B3}
\definecolor{IronGray}{HTML}{4D4D4D}
\definecolor{ScotsRose}{HTML}{FF1447}
\definecolor{GoldThread}{HTML}{FFAD00}
\definecolor{GreenThread}{HTML}{12DC00}
\definecolor{TealThread}{HTML}{00CC7A}
\definecolor{BlueThread}{HTML}{0023A3}
\definecolor{HighlandsSkyBlue}{HTML}{00C1D0}
\definecolor{LeonieGreen}{HTML}{00883A}
\definecolor{LeonieBlue}{HTML}{00329D}

\newcommand{\cmark}{\color{LeonieGreen}\ding{51}}%
\newcommand{\xmark}{\color{CarnegieRed}\ding{55}}%
\newcommand{\wug}{\color{LeonieBlue}\textbf{X}\color{black}}%

\newcommand{\affil}[1]{$^\textnormal{\textcolor{gray}{#1}}$}

\title{Verbing Weirds Language (Models): Evaluation of English Zero-Derivation in Five LLMs}

\name{\begin{tabular}{c}David R. Mortensen, \affil{1} Valentina Izrailevitch, \affil{2}Yunze Xiao, \affil{1}\\ Hinrich Schütze, \affil{3} Leonie Weissweiler\affil{3}\end{tabular}} 

\address{\affil{1}Carnegie Mellon University, \affil{2}TU Munich, \affil{3} LMU Munich \& MCML}

\abstract{
  Lexical-syntactic flexibility, in the form of conversion (or zero-derivation) is a hallmark of English morphology. In conversion, a word with one part of speech is placed in a non-prototypical context, where it is coerced to behave as if it had a different part of speech. However, while this process affects a large part of the English lexicon, little work has been done to establish the degree to which language models capture this type of generalization. This paper reports the first study on the behavior of large language models with reference to conversion. We design a task for testing lexical-syntactic flexibility---the degree to which models can generalize over words in a construction with a non-prototypical part of speech. This task is situated within a natural language inference paradigm. We test the abilities of five language models---two proprietary models (GPT-3.5 and GPT-4), three open-source models (Mistral 7B, Falcon 40B, and Llama 2 70B). We find that GPT-4 performs best on the task, followed by GPT-3.5, but that the open source language models are also able to perform it and that the 7B parameter Mistral displays as little difference between its baseline performance on the natural language inference task and the non-prototypical syntactic category task, as the massive GPT-4.
 \\ \newline \Keywords{morphology, syntax, large language models, open source} }

\begin{document}

\maketitleabstract

\input{sections/1-introduction.tex}
\input{sections/2-related_work.tex}
\input{sections/3-methods.tex}

\input{sections/4-results.tex}

\input{sections/5-discussion.tex}
\input{sections/6-conclusion.tex}

% \nocite{*}
\nocite{*}
\section{Bibliographical References}\label{sec:reference}

\bibliographystyle{lrec-coling2024-natbib}
\bibliography{lrec-coling2024-example}

\section{Language Resource References}
\label{lr:ref}
\bibliographystylelanguageresource{lrec-coling2024-natbib}

\end{document}

%% file: sections/1-introduction.tex
\begin{figure*}[b!]
    \centering
    \includegraphics[width=1.0\linewidth]{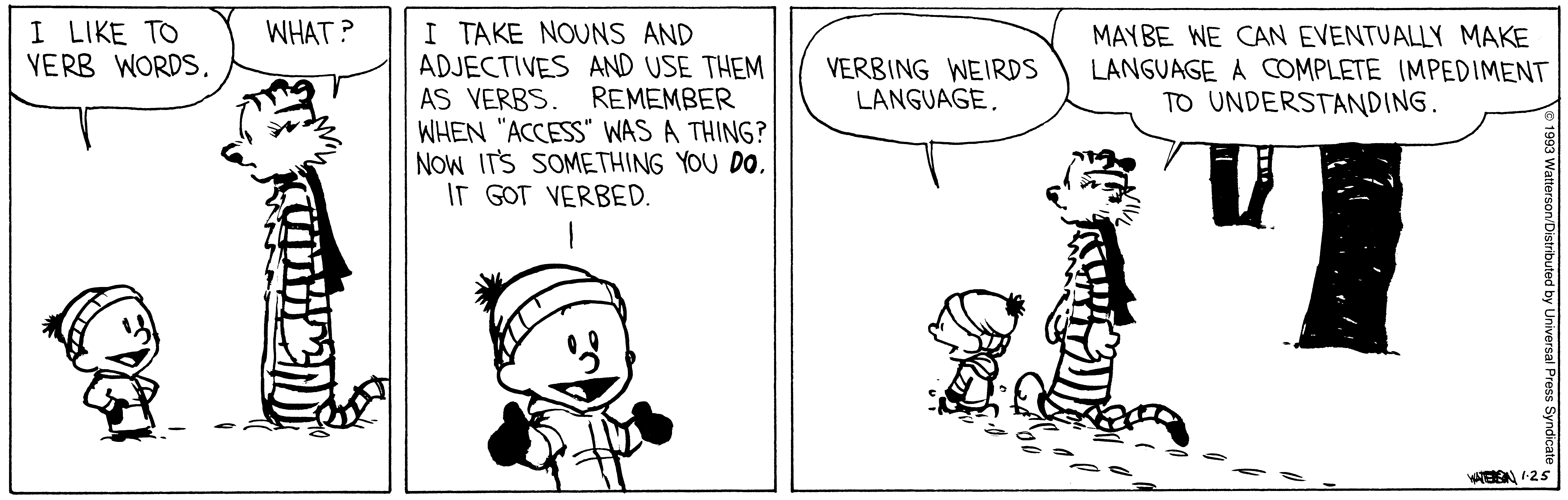}
    \caption{Calvin and Hobbes \textcopyright 1993 Watterson. Reprinted with permission of Andrews McMeel Syndication. All rights reserved.}
    \label{fig:comic}
\end{figure*}

\section{Introduction}
\label{sec:introduction}

English displays a relatively high degree of flexibility regarding the syntactic category of lexical items. This tendency is so pervasive that it is even commented on in popular culture, as in the American comic strip Calvin and Hobbes, in which one of the main characters proclaims ``I like to verb words'' (Figure~\ref{fig:comic}). In this case \textit{verb}---whose prototypical function is as a noun---functions as a verb. It is coerced into acting as a verb (specifically, an infinitive) by being placed as the head of a verb phrase, 
a position a noun could never occupy. In fact, some linguists have quipped, in only slightly hyperbolic fashion, that---in English---you can verb anything.

Other examples include:
\begin{description}[nosep]
\item[Adjective] His hair has begun to \textit{gray}.
\item[Mass Noun] If you don't want to \textit{water} the plants, please \textit{coffee} the graduate students instead.
\item[Count Noun] The fascist tried to \textit{knife} me in the back.
\end{description}
English also allows adjectives and verbs to be \textsc{converted} (zero-derived) into nouns:
\begin{description}[nosep]
\item[Intransitive Verb] I think I'll go for a \textit{swim}.
\item[Transitive Verb] I sustained a direct \textit{hit}.
\item[Adjective] She's got lots of \textit{green} but she's not spending any of it on me.
\end{description}
Note that, while \textit{swim}, \textit{hit}, and \textit{green} are treated as nouns (as well as verbs or adjectives) by lexicographers, they are etymologically verbs and adjectives.

There are three different linguistic approaches to this phenomenon. One is to say that conversion (or zero-derivation) is like any other morphological process (except that no overt affix, stress shift, or other formal change is evident) \citep{marchand1969categories}. The principal evidence for this, as pointed out by \citet{beard2017derivation}, is that conversion is subject to blocking effects: if a word of the same meaning as a potential converted word already exists in the lexicon (e.g., due to derivation), it will be used instead of the converted word. The appeal of this approach is that it accounts for the many different semantic relationships that can exist between bases and zero-derived words with the same parts of speech. Contrariwise, it has been proposed that conversion is not morphological at all---that it is, effectively, coining new words \citep{lieber1980organization}. Our study is conducted in the spirit of a third approach, namely that English, Dutch, and other, similar, languages allow flexibility with regard to syntactic categories (parts of speech) when this is (1) allowed by the context semantically and (2) required by the context syntactically \citep{clark1979nouns}. We do not attempt to implement these three approaches computationally, or to distinguish them empirically, but methodologically we manipulate part of speech by manipulating aspects of the syntactic and semantic context, inspired by insights of \citet{clark1979nouns}.

In this study, we evaluate five large language models of varying sizes, based on their ability to make inferences that require reasoning about words used in non-prototypical grammatical contexts. We investigate four major hypotheses:
\begin{itemize}
\setlength\itemsep{.5em}
    \item performance on the task with prototypical parts of speech is better than with non-prototypical parts of speech
    \item non-prototypical parts of speech are associated with better performance than nonce words
    \item correlation between performance in the prototypical, non-prototypical, and nonce conditions
    \item differences in model size account for differences in performance
\end{itemize}

Performance in the prototypical condition is---indeed---the best, but performance in the nonce and non-prototypical conditions are similar. The performance of each model was correlated across conditions. We find that the models vary greatly in their ability in this area, with the very large, commercial models performing best. However, we also show that the number of parameters alone does not predict the performance of models on this task. Instead, a good predictor is the performance of the models on a generic version of the task in which all words are used in prototypical ways. This suggests that most of the variance in the model scores is variance in the ability to perform the NLI task itself, and that the other differences are of lesser, but still significant, importance.

We make three main contributions:
\begin{itemize}
\setlength\itemsep{.5em}
\item A new methodology for investigating conversion in language models
\item A test set for systematically applying this methodology to arbitrary models
\item The demonstration that lexical-syntactic flexibility does not increase monotonically with model size
\end{itemize}

%%% Local Variables:
%%% mode: latex
%%% TeX-master: "../lrec-coling2024"
%%% TeX-engine: xetex
%%% End:

%% file: sections/2-related_work.tex
\section{Related Work}

Linguistic work on conversion in English dates back to \citet{sweet1981new}. It has been taken up sporadically by researchers since then \citep{marchand1969categories,clark1979nouns,lieber1980organization, kiparsky1982cyclic, don1993morphological, bauer2005approaches}. 
Much of the literature regarding conversion concerns whether conversion is due to a kind of zero-affixation \citep{marchand1969categories}, a process of coinage \citep{lieber1980organization}, or the flouting of syntactic category constraints when constrained (and allowed) by context \citep{clark1979nouns}. 
This paper assumes the position of \citeauthor{clark1979nouns}, namely that languages like English allow syntactic flexibility when licensed by semantics and required by the encompassing constructions. 
This is analogous, in some ways, to \citet{goldberg1995constructions}'s analysis of argument structure alternations, where the broader construction coerces, for example, intransitive verbs to function as transitive verbs (in the caused-motion construction). Similarly, we assume that constructional contexts coerce words to function as if they have a non-prototypical part of speech.

While this is the first study modeling conversion computationally, there is a growing body of work addressing related issues for a broader range of phenomena in derivational morphology and neologism. The most relevant to the current work are \citet{hofmann-etal-2021-superbizarre} and \citet{hofmann-etal-2020-dagobert}, which address derivational morphology in the context of older BERT-like language models (but not contemporary LLMs). Factors in the emergence of new words have been elucidated by \citet{ryskina2020where}.

%%% Local Variables:
%%% mode: latex
%%% TeX-master: "../lrec-coling2024"
%%% TeX-engine: xetex
%%% End:

%% file: sections/3-methods.tex
\section{Methodology}
\label{sec:methodology}

We sought to design a task that would test whether words from non-prototypical syntactic categories---converted words and nonce words---affect the ability of text-in-text-out language models to make pragmatic generalizations.

\subsection{Materials}

We created a set of 3,069 prompts based on the frames in Table~\ref{tab:frames} and five word lists (Table~\ref{tab:word-lists}).
\begin{table}[h]
  \centering
  \begin{tblr}{colspec={lrl}}
    \toprule
    Part of Speech & Number & Example\\
    \midrule
    transitive verbs & 42 & bamboozle \\
    intransitive verbs & 42 & deign \\
    mass nouns & 51 & music\\
    count nouns & 79 & professor\\
    nonce words & 49 & theord\\
    \bottomrule
  \end{tblr}
  \caption{Word lists derived from UniMorph and Unipseudo \cite{unipseudo} and verified manually}
  \label{tab:word-lists}
\end{table}

Nonce words were generated with Unipseudo \citep{unipseudo} based on a list of the 59 most frequent mono-morphemic nouns verbs in English with length 6 (as listed in UniMorph). This list was manually culled to remove words that were (1) too similar to or (2) too distant from any known English words.
All lexical sets were manually curated by a native-speaker linguist.

\begin{table*}[tbh!]
  \centering\footnotesize
  \input{tables/frames.tex}

  \caption{Frames used for generating prompts}
  \label{tab:frames}
\end{table*}

The frames and the wordlists were combined according to principled criteria, yielding 3,069 items. Prompts reflect the format of the following example:
\begin{quote}
    If I asked you to \textbf{day} it, do I want you to \textbf{day} it?
\end{quote}

\subsection{Experiments}

The prompts were presented to five models, two closed models (GPT-3.5 and GPT-4) and three open models of varying sizes (Mistral 7B, Falcon 40B, and Llama 2 70B). The closed models were prompted via the OpenAI API. The open models were all evaluated (without quantization) using \texttt{vLLM} on a cluster node with 4 A6000 GPUs. The prompts described above were presented to the models with the suffix, `` Answer with one word.'' Answers were automatically identified using regular expressions.
Responses starting with ``yes'', ``yeah'', ``sure'', ``correct'', ``right'', and ``true'' were coded as affirmative and those starting with ``no'', ``nope'', ``wrong'', ``incorrect'', and ``false'' were coded as negative. Other responses were treated as ``null''. For each model, four runs (of 3,069 prompts) were made.

%%% Local Variables:
%%% mode: latex
%%% TeX-master: "../lrec-coling2024"
%%% TeX-engine: xetex
%%% End:

%% file: tables/frames.tex
\begin{tabular}{lll}
  \toprule
  Subtask & Prompt & Intended\\
  \midrule
\multirow{6}{*}{transitive} &
If I asked you to \wug{} it, do I want you to  $\left\{\begin{array}{l} \emptyset \\ \text{not}\end{array}\right\}$ \wug{} it? & $\begin{array}{l} \text{\cmark} \\ \text{\xmark}\end{array}$
\\
&If I asked you $\left\{\begin{array}{l}\text{not to} \\ \text{to not}\end{array}\right\}$ \wug{} it, do I want you to $\left\{\begin{array}{l} \emptyset \\ \text{not}\end{array}\right\}$ \wug{} it?& $\begin{array}{l}\text{\xmark} \\ \text{\cmark}\end{array}$
\\
&If I say, “Don’t \wug{} me,” am I asking you to $\left\{\begin{array}{l}\text{not} \\ \end{array}\right\}$ \wug{} me?& $\begin{array}{l}\text{\cmark} \\ \text{\xmark}\end{array}$
\\
\midrule
\multirow{4}{*}{intransitive} & If I$\left\{\begin{array}{l} \emptyset \\ \text{don't}\end{array}\right\}$\wug{} daily, do I \wug{} every day? &$\begin{array}{l}\text{\cmark} \\ \text{\xmark}\end{array}$\\
 & If I$\left\{\begin{array}{l} \text{tried}\\ \text{did not try}\end{array}\right\}$ to \wug{}, did I attempt to \wug{}? & $\begin{array}{l}\text{\cmark} \\ \text{\xmark}\end{array}$\\
\midrule
\multirow{4}{*}{count} &  If I like this \wug{} more than the other one, do I prefer $\left\{\begin{array}{l} \text{this \wug{} to the other one?}\\ \text{the other \wug{} to this one?}\end{array}\right\}$ & $\begin{array}{l}\text{\cmark} \\ \text{\xmark}\end{array}$ \\
 &  If I like this \wug{} less than the other one, do I prefer $\left\{\begin{array}{l} \text{this \wug{} to the other one?}\\ \text{the other \wug{} to this one?}\end{array}\right\}$ & $\begin{array}{l}\text{\xmark} \\ \text{\cmark}\end{array}$ \\
\midrule
\multirow{4}{*}{mass}&
I prefer less \wug{}. Do I prefer $\left\{\begin{array}{l}\text{less} \\ \text{more}\end{array}\right\}$ \wug{}? & $\begin{array}{l}\text{\cmark} \\ \text{\xmark}\end{array}$\\
&I prefer more \wug{}. Do I prefer $\left\{\begin{array}{l}\text{less} \\ \text{more}\end{array}\right\}$ \wug{}? & $\begin{array}{l}\text{\xmark} \\ \text{\cmark}\end{array}$\\
\bottomrule
\end{tabular}

%% file: sections/4-results.tex
\section{Results}
\label{sec:results}

\begin{figure}
    \centering
    \includegraphics[width=\linewidth]{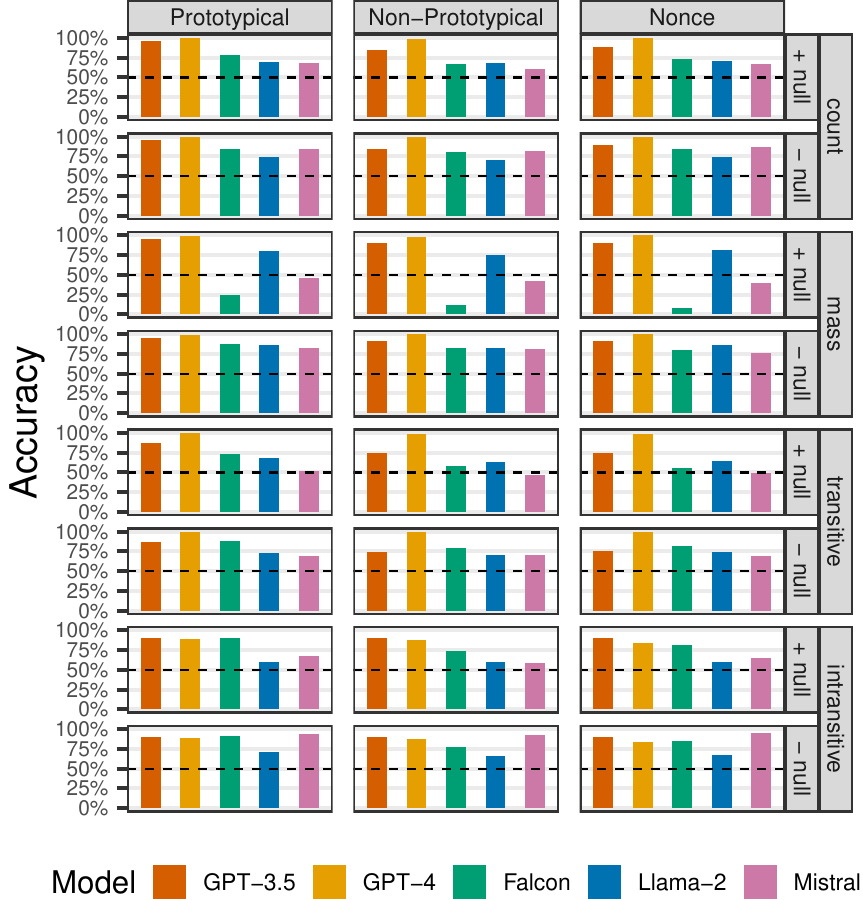}
    \caption{Average accuracy grouped by model and typicality (p- for ``prototypical,'' np- for ``non-prototypical,'' and no- for ``nonce'')}
    \label{fig:results}
\end{figure}

Overall results are shown in Figure~\ref{fig:results}.  GPT-4 achieves almost perfect results (maximal lexical-syntactic flexibility) across all categories (count noun frames, mass noun frames, and transitive verb frames). The exception is the intransitive verb frames, where its performance, when nulls are removed, is worse than that of Mistral 7B. GPT-3.5 is consistently worse than GPT-4 but is, on balance, a stronger performer than the open-source models. Falcon 40B performs better on the metric than the other open-source models, on the prototypical condition. The glaring exception is the mass noun frames, where Falcon generated mostly non-sequiturs rather than yes/no responses. When all responses are considered, Mistral is a relatively weak performer. However, when null responses are filtered out, Mistral appears to display greater flexibility than the other open-source models.

In order to separate the ability to carry out the natural language inference task from lexical-syntactic flexibility, we analyzed the difference between the average accuracy on the prototypical condition (expected part of speech) and the non-prototypical condition (unexpected part of speech). These results are shown in Table~\ref{tab:contrasting-difference}. The inaccuracy associated with non-prototypical contexts was substantially higher for GPT-3.5 and Falcon 40B, but was relatively low for Llama 2 70B and was minimal for Mistral 7B and GPT-4.

\begin{table}[tbh]
  \centering
  \input{tables/contrasting_difference.tex}
  \caption{Difference in average accuracy between the prototypical and non-prototypical condition, both with all records (left, null treated as negative) and with only non-null records (right)}
  \label{tab:contrasting-difference}
\end{table}

%%% Local Variables:
%%% mode: latex
%%% TeX-master: "../lrec-coling2024"
%%% TeX-engine: xetex
%%% End:

%% file: tables/contrasting_difference.tex
\begin{tblr}{
    colspec={lrr},
  }
\toprule
model & with nulls & without nulls \\
\midrule
gpt-3.5 & 0.08 & 0.08 \\
gpt-4 & 0.01 & 0.01 \\
\midrule
falcon & 0.15 & 0.07 \\
llama2 & 0.03 & 0.03 \\
mistral & 0.07 & 0.01 \\
\bottomrule
\end{tblr}

%%% Local Variables:
%%% mode: latex
%%% TeX-engine: xetex
%%% TeX-master: "../"
%%% End:

%% file: sections/5-discussion.tex
\section{Discussion}
\label{sec:discussion}
In order to understand which factors most contributed to performance on the syntactic flexibility task, we fitted a Logistic Regression to these results using the \texttt{Logit} function from the Python \texttt{statsmodels} library. The features were prototypical part of speech, model type, (proto)typicality of the filler given the frame, and whether the answer was yes, no, or ``null''. It showed all factors to be significant predictors of correctness ($p < 0.01$), with answer type (yes, no, or null) as the strongest predictor. This is likely because Mistral and Falcon often fail to give correct responses by generating answers that are coded neither ``yes'' nor ``no''. This is associated with a confounder (these frames, with two sentences, often elicited null responses). Controlling for all of the other factors, model type is also predictive, with GPT-3.5 and GPT-4 most associated with correct responses and Llama 2 least associated with correct responses.

Returning to our hypotheses in Table \ref{tab:hypotheses}, we find that the models do, almost without exception, perform better under the prototypical condition than the non-prototypical condition. This suggests that conversion is more challenging than the use of unconverted words. However, non-prototypical performance is not significantly different from nonce performance, suggesting that the models are treating converted words as, essentially, nonce words. Scores for all three conditions are generally correlated with one another---models that are good at using words in a prototypical way are also good at using them in non-prototypical ways. Perhaps most surprising, though, is the fact that model size was not a good predictor of performance. The largest of the open models (Llama 2 70B) was in some ways the weakest performer. Mistral 7B, was the smallest, but held its own against Falcon 40B and even the much larger GPT models, and least in certain subtasks. This was true in spite of the fact that Mistral and Falcon were generally worse at following instructions.

Investigating the differences between the models in detail, it is clear that GPT-4 displays, far and away, the best scores on our lexical-syntactic flexibility task. It might be tempting to attribute this difference to the number of parameters in the model (since GPT-4 is believed to be a Mixture of Experts of several large models). However, it is not the case that this kind of generalization is necessarily simply a function of model size: the best-performing open-source model, when all else is held equal, is also the smallest (Mistral 7B). The largest of the open-source models---Llama 2 70B---is consistently mediocre in its performance on this task. And Falcon 40B, which is almost six times the size of Mistral, shows impressive abilities at the natural language inference task but lackluster performance at lexical-syntactic flexibility.

Mistral and Falcon's scores are hurt by the fact that they frequently general null responses, particularly in response to frames eliciting mass nouns (Falcon and Mistral) and transitive verbs (Mistral). The causal mechanism, in these cases, is unclear. The mass noun frames all consist of two sentences: ``I prefer more/less X. Do I prefer more/less X?'' The other frames have one sentence each. The intransitive frames require the model to reason about semantically related words (though the count noun subtask does as well). The generation of null responses may be due either to these superficial factors or to more basic differences in model behavior with respect to frames that call for mass nouns or intransitive verbs.

GPT-4 also displays a dip in performance on the intransitive subtask, not because it is generating null responses but because it is generating incorrect yes/no responses in a non-trivial number of cases. Again, because the sets of frames are so small and lacking in diversity, it is not possible to construct a valid causal explanation to account for the fact that the models perform differently on some subtasks than others. What is clear, though, is that there are significant differences between the models and these differences are in some way correlated with the subtasks defined here.

%%% Local Variables:
%%% mode: latex
%%% TeX-master: "../lrec-coling2024"
%%% TeX-engine: xetex
%%% End:

%% file: sections/6-conclusion.tex
\begin{table*}[t]
  \centering
  \input{tables/hypotheses.tex}
  \caption{Findings with regard to each major hypothesis}
  \label{tab:hypotheses}
\end{table*}

\section{Conclusion}
\label{sec:future-directions}

We have introduced the first experiment testing the lexical-syntactic flexibility of LLMs, finding that language models are challenged by words in syntactically non-prototypical context (when compared to words in syntactically prototypical contexts). However, we did not find that words in syntactically non-prototypical contexts presented challenges to the models that nonce words did not. As we posited, there is a correlation between performance on prototypical and non-prototypical items and the model type was a significant predictor of performance. However, contrary to expectations, the model size was not a good predictor of lexical-syntactic flexibility. The findings are summarized in Table~\ref{tab:hypotheses}.

With this foundation in place, we plan to investigate lexical-syntactic flexibility more systematically by using a much larger number of frames for each subtask and by testing a larger set of (open and proprietary) models. Now that truly open models like Olmo \citep{groeneveld2024olmo} are available, it is possible to know, more precisely, what words have been seen by the model and in what contexts. This will allow us to state unambiguously when models are generalizing old vocabulary to new contexts and when they are directly recapitulating what they have seen in their training data.

%% file: tables/hypotheses.tex
\begin{tblr}{
    colspec={Xl},
    cell{2,4}{2} = {fg=CarnegieRed},
    cell{1,3}{2} = {fg=GreenThread},
    cell{1}{2} = {fg=black},
    row{1} = {c},
  }
  \toprule
  Hypothesis & Finding\\
  \midrule
  prototypical performance $>$ non-prototypical performance & \cmark{} Supported\\
  non-prototypical performance $>$ nonce performance & \xmark{} Not supported\\
  Correlation between prototypical, non-prototypical, nonce performance & \cmark{} Supported\\
  Difference between model size accounts for difference in performance & \xmark{} Not supported\\
  \bottomrule
\end{tblr}  

%%% Local Variables:
%%% mode: latex
%%% TeX-master: t
%%% End:

%% file: lrec-coling2024.bbl
\begin{thebibliography}{18}
\expandafter\ifx\csname natexlab\endcsname\relax\def\natexlab#1{#1}\fi

\bibitem[{Bauer and Hern{\'a}ndez(2005)}]{bauer2005approaches}
Laurie Bauer and Salvador~Valera Hern{\'a}ndez. 2005.
\newblock \emph{Approaches to conversion/zero-derivation}.
\newblock Waxmann Verlag.

\bibitem[{Bauer and Valera(2005)}]{bauer2005conversion}
Laurie Bauer and Salvador Valera. 2005.
\newblock Conversion or zero-derivation: An introduction.
\newblock \emph{Approaches to conversion/zero-derivation}, pages 7--17.

\bibitem[{Beard(2017)}]{beard2017derivation}
Robert Beard. 2017.
\newblock \href {https://doi.org/https://doi.org/10.1002/9781405166348.ch2} {\emph{Derivation}}, chapter~2. John Wiley \& Sons, Ltd.

\bibitem[{Clark and Clark(1979)}]{clark1979nouns}
Eve~V Clark and Herbert~H Clark. 1979.
\newblock When nouns surface as verbs.
\newblock \emph{Language}, pages 767--811.

\bibitem[{Don(1993)}]{don1993morphological}
Jan Don. 1993.
\newblock \emph{Morphological Conversion}.
\newblock Ph.D. thesis, University of Utrecht.

\bibitem[{Francis et~al.(2021)Francis, Rabinovich, Samir, Mortensen, and Stevenson}]{francis-etal-2021-quantifying}
David Francis, Ella Rabinovich, Farhan Samir, David Mortensen, and Suzanne Stevenson. 2021.
\newblock \href {https://doi.org/10.1162/tacl_a_00441} {Quantifying cognitive factors in lexical decline}.
\newblock \emph{Transactions of the Association for Computational Linguistics}, 9:1529--1545.

\bibitem[{Goldberg(1995)}]{goldberg1995constructions}
Adele~E. Goldberg. 1995.
\newblock \emph{Constructions: A Construction Grammar Approach to Argument Structure}.
\newblock University of Chicago Press, Chicago, IL.

\bibitem[{Groeneveld et~al.(2024)Groeneveld, Beltagy, Walsh, Bhagia, Kinney, Tafjord, Jha, Ivison, Magnusson, Wang, Arora, Atkinson, Authur, Chandu, Cohan, Dumas, Elazar, Gu, Hessel, Khot, Merrill, Morrison, Muennighoff, Naik, Nam, Peters, Pyatkin, Ravichander, Schwenk, Shah, Smith, Strubell, Subramani, Wortsman, Dasigi, Lambert, Richardson, Zettlemoyer, Dodge, Lo, Soldaini, Smith, and Hajishirzi}]{groeneveld2024olmo}
Dirk Groeneveld, Iz~Beltagy, Pete Walsh, Akshita Bhagia, Rodney Kinney, Oyvind Tafjord, Ananya~Harsh Jha, Hamish Ivison, Ian Magnusson, Yizhong Wang, Shane Arora, David Atkinson, Russell Authur, Khyathi~Raghavi Chandu, Arman Cohan, Jennifer Dumas, Yanai Elazar, Yuling Gu, Jack Hessel, Tushar Khot, William Merrill, Jacob Morrison, Niklas Muennighoff, Aakanksha Naik, Crystal Nam, Matthew~E. Peters, Valentina Pyatkin, Abhilasha Ravichander, Dustin Schwenk, Saurabh Shah, Will Smith, Emma Strubell, Nishant Subramani, Mitchell Wortsman, Pradeep Dasigi, Nathan Lambert, Kyle Richardson, Luke Zettlemoyer, Jesse Dodge, Kyle Lo, Luca Soldaini, Noah~A. Smith, and Hannaneh Hajishirzi. 2024.
\newblock \href {http://arxiv.org/abs/2402.00838} {Olmo: Accelerating the science of language models}.

\bibitem[{Hofmann et~al.(2020{\natexlab{a}})Hofmann, Pierrehumbert, and Sch{\"u}tze}]{hofmann-etal-2020-dagobert}
Valentin Hofmann, Janet Pierrehumbert, and Hinrich Sch{\"u}tze. 2020{\natexlab{a}}.
\newblock \href {https://doi.org/10.18653/v1/2020.emnlp-main.316} {{D}ago{BERT}: {G}enerating derivational morphology with a pretrained language model}.
\newblock In \emph{Proceedings of the 2020 Conference on Empirical Methods in Natural Language Processing (EMNLP)}, pages 3848--3861, Online. Association for Computational Linguistics.

\bibitem[{Hofmann et~al.(2021)Hofmann, Pierrehumbert, and Sch{\"u}tze}]{hofmann-etal-2021-superbizarre}
Valentin Hofmann, Janet Pierrehumbert, and Hinrich Sch{\"u}tze. 2021.
\newblock \href {https://doi.org/10.18653/v1/2021.acl-long.279} {Superbizarre is not superb: Derivational morphology improves {BERT}{'}s interpretation of complex words}.
\newblock In \emph{Proceedings of the 59th Annual Meeting of the Association for Computational Linguistics and the 11th International Joint Conference on Natural Language Processing (Volume 1: Long Papers)}, pages 3594--3608, Online. Association for Computational Linguistics.

\bibitem[{Hofmann et~al.(2020{\natexlab{b}})Hofmann, Sch{\"u}tze, and Pierrehumbert}]{hofmann-etal-2020-graph}
Valentin Hofmann, Hinrich Sch{\"u}tze, and Janet Pierrehumbert. 2020{\natexlab{b}}.
\newblock \href {https://doi.org/10.18653/v1/2020.acl-main.106} {A graph auto-encoder model of derivational morphology}.
\newblock In \emph{Proceedings of the 58th Annual Meeting of the Association for Computational Linguistics}, pages 1127--1138, Online. Association for Computational Linguistics.

\bibitem[{Kiparsky(1982)}]{kiparsky1982cyclic}
Paul Kiparsky. 1982.
\newblock From cyclic phonology to lexical phonology.
\newblock In H.~van~der Hulst and N.~Smith, editors, \emph{The Structure of Phonological Representations}, pages 131--175. Foris, Dordrecht.

\bibitem[{Lieber(1980)}]{lieber1980organization}
Rochelle Lieber. 1980.
\newblock \emph{On the organization of the lexicon}.
\newblock Ph.D. thesis, University of New Hampshire.

\bibitem[{Marchand(1969)}]{marchand1969categories}
Hans Marchand. 1969.
\newblock \emph{The Categories and Types of Present-Day English Word-Formation}.
\newblock C. H. Beck, M\"unchen.

\bibitem[{New et~al.(in press)New, Pallier, Bourgin, and Barra}]{unipseudo}
Boris New, Christophe Pallier, Jessica Bourgin, and Julien Barra. in press.
\newblock Unipseudo.
\newblock http://www.lexique.org/shiny/unipseudo/.
\newblock Accessed: 19.10.2023.

\bibitem[{Ryskina et~al.(2020)Ryskina, Rabinovich, Berg-Kirkpatrick, Mortensen, and Tsvetkov}]{ryskina2020where}
Maria Ryskina, Ella Rabinovich, Taylor Berg-Kirkpatrick, David~R. Mortensen, and Yulia Tsvetkov. 2020.
\newblock Where new words are born: Distributional semantic analysis of neologisms and their semantic neighborhoods.
\newblock In \emph{Proceedings of the Society for Computation in Linguistics}, volume~3.

\bibitem[{Sweet(1891)}]{sweet1981new}
Henry Sweet. 1891.
\newblock \emph{A new {English} grammar, logical and historical. Part I. Introduction, phonology, and accidence}.
\newblock Clarendon Press, Oxford.

\bibitem[{Weissweiler et~al.(2023)Weissweiler, Hofmann, Kantharuban, Cai, Dutt, Hengle, Kabra, Kulkarni, Vijayakumar, Yu, Sch\"utze, Oflazer, and Mortensen}]{weissweiler2023counting}
Leonie Weissweiler, Valentin Hofmann, Anjali Kantharuban, Anna Cai, Ritan Dutt, Amey Hengle, Anubha Kabra, Atharva Kulkarni, Abhishek Vijayakumar, Haofei Yu, Hinrich Sch\"utze, Kemal Oflazer, and David~R. Mortensen. 2023.
\newblock Counting the bugs in {ChatGPT}'s wugs: A multilingual investigation into the morphological capabilities of a large language model.
\newblock In \emph{Proceedings of the 2023 Conference on Empirical Methods in Natural Language Processing (EMNLP)}, Online. Association for Computational Linguistics.

\end{thebibliography}
